\def\year{2019}\relax
\documentclass[letterpaper]{article} 
\usepackage{aaai19}  
\usepackage{times}  
\usepackage{helvet}  
\usepackage{courier}  
\usepackage{url}  
\usepackage{graphicx}  
\usepackage{mathtools}
\usepackage{subfigure}
\usepackage{amsmath}
\usepackage{amsfonts}

\begin{document}
%
\title{Neural Allocentric Intuitive Physics Prediction from Real Videos}
\author{Zhihua Wang, Stefano Rosa, Yishu Miao, Zihang Lai, Linhai Xie, Andrew Markham, Niki Trigoni\\
Department of Computer Science, University of Oxford}
\maketitle
\begin{abstract}
Humans are able to make rich predictions about the future dynamics of physical objects from a glance. 
On the other hand, most existing computer vision approaches require strong assumptions about the underlying system, ad-hoc modeling, or annotated datasets, to carry out even simple predictions.
To tackle this gap, we propose a new perspective on the problem of learning intuitive physics that is inspired by the spatial memory representation of objects and spaces in human brains, in particular the co-existence of egocentric and allocentric spatial representations.
We present a generic framework that learns a layered representation of the physical world, using a cascade of invertible modules.
In this framework, real images are first converted to a synthetic domain representation that reduces complexity arising from lighting and texture.
Then, an allocentric viewpoint transformer removes viewpoint complexity by projecting images to a canonical view.
Finally, a novel \emph{Recurrent Latent Variation Network} (RLVN) architecture learns the dynamics of the objects interacting with the environment and predicts future motion, leveraging the availability of unlimited synthetic simulations. 
Predicted frames are then projected back to the original camera view and translated back to the real world domain.
Experimental results show the ability of the framework to consistently and accurately predict several frames in the future and the ability to adapt to real images.
\end{abstract}

\section{Introduction}

\begin{figure*}[h]
  \centering
  \includegraphics[width=0.8\textwidth]{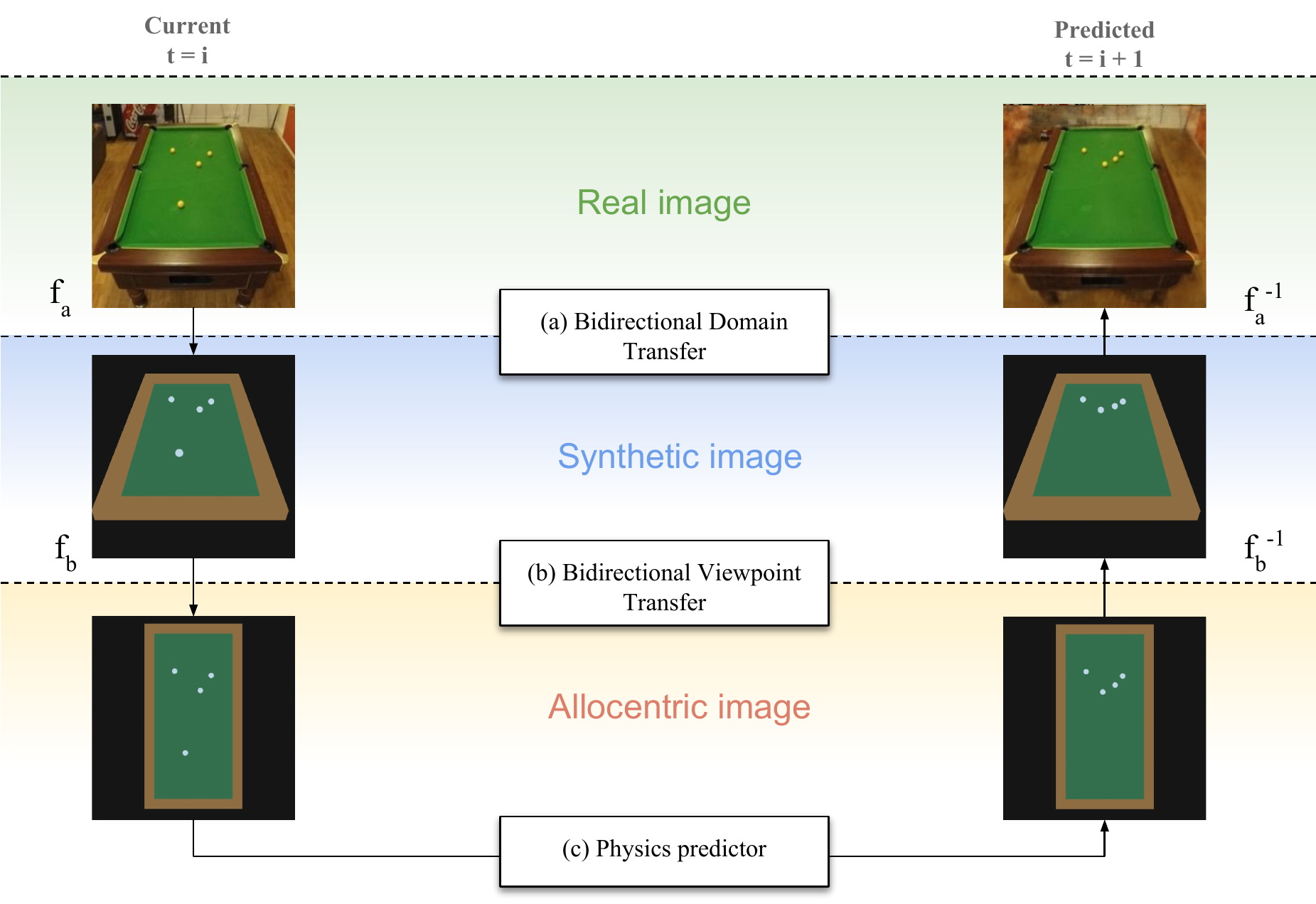}
  \caption{Overview of the framework.}
  \label{fig:architecture}
\end{figure*}

Humans have certain expectations about the physical world and learn to estimate mass and velocity of objects at an early stage of development through observation \cite{spelke2007core}.
The problem of learning the intuitive dynamics of objects from data is often referred to as \emph{learning intuitive physics}.
Applications of intuitive physics are especially promising in the field of robotics, including manipulation, navigation and co-working scenarios.
A robot equipped with intuitive physics understanding is able to navigate the environment and perform nuanced actions, such as carrying a cup of coffee without spilling it, catching a falling tool, and so on. 

At a more generic level, physical understanding is a core domain of human knowledge and amongst the earliest topics in artificial intelligence. However, devising systems for physical reasoning that are able to learn from a few real unlabeled images is still an open problem.
One major challenge is the limited amount of available data collected from the real world, which are generally sparse and lack of annotations. In addition, the quality of collected real world dataset is usually impacted by factors such as illumination, occlusion and perspective. Normally, not only can objects move within the scene, but the observer themselves can be shifting as well, which further complicates the predictions. Therefore, in order to learn intuitive physics with machines, we attempt to seek the inspirations from human brains.

At the core of physical reasoning lies a spatial representation of objects and the environment. In the human brain, spatial representation is necessary for navigating through known or unknown environments, locating objects and interacting with them. 
An idea was introduced in the mid 20th century (Figure \ref{fig:spatialmemory}) that information in these \emph{cognitive maps} is represented with two types of \emph{frames}\footnote{The frame in this context bears a more relaxed meaning compared to geometric reference frames.}.
The \emph{egocentric frame} represents the information from the perspective of the observer in the environment, while the \emph{allocentric frame} represents the information about the spatial relationship of objects relative to each other.
The egocentric representation focuses on subject-to-object relationships, which is view-dependent and generally believed to be learnt first during early development, while the allocentric frame is based on world-based (global) coordinates and is believed to be acquired later in life \cite{colombo2017egocentric}. 
These systems have already informed some approaches in the computer vision community for tasks such as mapping \cite{henriques2018mapnet}. 

Two models of spatial cognition have been proposed. In the \emph{two-system model}, an allocentric representation of spatial relationships between objects is stored in the long-term memory, while a self-reference system keeps track of egocentric relations to each object.
In the \emph{three-system model}, a dynamic egocentric system stores relationships between the observer and each object in its neighborhood, while a second system maintains allocentric representation in the long-term memory and a third one stores visual snapshots of the environment at different times \cite{avraamides2008multiple}.

By taking inspiration from these two models, we propose a framework (Figure \ref{fig:architecture}) for learning the intuitive dynamics of objects interacting in the environment which starts from ego-centric observations, warps these into an allocentric view of the scene, and learns the dynamics using a recurrent latent variation model. The predictor makes predictions about future observations in the allocentric frame, that are then translated back to the ego-centric view of the observer \ref{fig:architecture}.

In order to learn to predict future frames, we use a realistic physics simulator (the framework is data-based and hence agnostic to the simulator used) to generate synthetic observations. The two advantages of using synthetic data are the possibility to train the egocentric to allocentric warping module in a supervised way and the abstraction of real-world images to synthetic images, which removes most lighting, color, texture artifacts that are irrelevant to the task of predicting object dynamics.  

The main contributions of this work are the following: a neuro-inspired framework that tackles the problem of predicting the future state of objects from real video inputs with arbitrary viewpoints by projecting the scene to and from an allocentric representation; a Recurrent Latent Variation Network (RLVN), based on Convolutional LSTM networks (ConvLSTMs) \cite{xingjian2015convolutional}, able to predict the dynamics of objects interacting among themselves and with the sourrounding environment, with improved long-term prediction capability compared to other state-of-the-art approaches. 
In particular, we demonstrate the performance of this novel framework on the problem of predicting the motion of billiard balls.

\section{Related Work}
Prediction of the dynamics of physical objects lies at the intersection between two bodies of work: future frame prediction, which focuses on global frames, and learning intuitive physics, which often focuses on object-based representations.

\paragraph{Learning intuitive physics}
In an early work \cite{wu2015galileo} the authors first proposed to use deep generative models for learning the effect of gravity and friction on rolling objects by inverting a physics engine, in order to estimate the dynamics from observations.
Deep neural networks have been later used for predicting the stability of tower blocks \cite{lerer2016learning}, motion of billiard balls \cite{Fragkiadaki16} and other object dynamics \cite{mottaghi2016newtonian}. 


Differentiable physics engines have been proposed in \cite{chang2016compositional}.

Applications of intuitive physics to robotics have been recently explored in \cite{byravan2017se3,3dphysnet}, 
for predicting rigid and non-rigid body motion of objects subjected to forces, or in \cite{li2017visual} for stability prediction in stacking blocks.

Recently, \cite{vda17} proposed to decouple the prediction problem by learning an abstract physical representation of the world with a perception network, and using the physical representation as input to a physics engine and a rendering engine in order to generate visual data, which can be then matched to the visual input. One advantage of such approaches is that it is able to generate very sharp predictions. However, a disadvantage is that different simulation engines and renderers are required for different tasks, which leads to the poor generalization ability. 


Interaction Networks \cite{battaglia2016interaction} model interactions combine a relational reasoning network and an object reasoning network to predict object dynamics in a similar fashion to simulators.  By adding the vector outputs of all object interactions, a global interaction vector is obtained, that is used together with object features to predict the future velocity of each object.
Visual Interaction Networks \cite{NIPS2017_7040} learn to predict future trajectories of objects in a physical system from video frames by jointly training a perceptual front-end based on convolutional networks and a dynamics predictor based on interaction networks. In \cite{Ehrhardt2017TakingVM}, the focus is on learning the motion of balls on non-homogeneous surfaces.

More recently, Relational-NEM \cite{van2018relational} proposes a compositional approach for unsupervised learning the dynamics of multiple bouncing balls. The approach is focused on interactions between multiple objects, thus an estimate of the boundary conditions of the problem is required (e.g., the number of existing objects). Moreover, a noise is injected into input images in order to help learning of object grouping; tuning of the injected noise can be an issue when dealing with small objects.
In addition, Pred-RNN proposes causal LSTM cells and Pred-RNN++ \cite{wang2018predrnn} addresses the problem of balancing long-term predictions with the induced difficulty in back-propagation with a Gradient Highway architecture, providing alternative routes for gradient flow.

In \cite{Bhattacharyya2018LongTermIB} the authors propose a context-base model for predicting image boundaries in future frames, and apply it to the problem of predicting the motion of billiard balls among other scenarios.

\paragraph{Predicting future video frames}
Among the first works on future frame prediction, \cite{MathieuCL15} proposed a CNN architecture with adversarial training. \cite{srivastava2015unsupervised} uses multi-layer LSTMs for unsupervised learning.
In \cite{xue2016visual}, the authors propose an architecture called Cross Convolutional Networks, that encodes image and motion information separately as feature maps and convolutional kernels, respectively.

Predictive Neural Network (PredNet) architectures \cite{prednet16} are inspired by the concept of predictive coding from a neuroscience perspective. These networks learn to predict future frames in a video sequence, with each layer in the network making local predictions and forwarding deviations from those predictions to successive network layers.

\section{Method}
The framework is composed of three modules: the domain transfer module works at the lower level and translates image appearence between the real world and a simplified synthetic domain; the egocentric to allocentric transformer works at an intermediate level to translate egocentric images to a canonical allocentric view; the phsyics predictor module works at the physical level and learns the properties of the objects and the scene. An overview of the framework is shown in Figure \ref{fig:architecture}.

We first describe our physics predictor network, then we show how to go from simulated data to real world data with the domain transfer module and the allocentric transformer.





\subsection{Real-world to Synthetic Data Domain Transfer} 

\begin{figure}[t]
  \centering
  \includegraphics[width=\columnwidth]{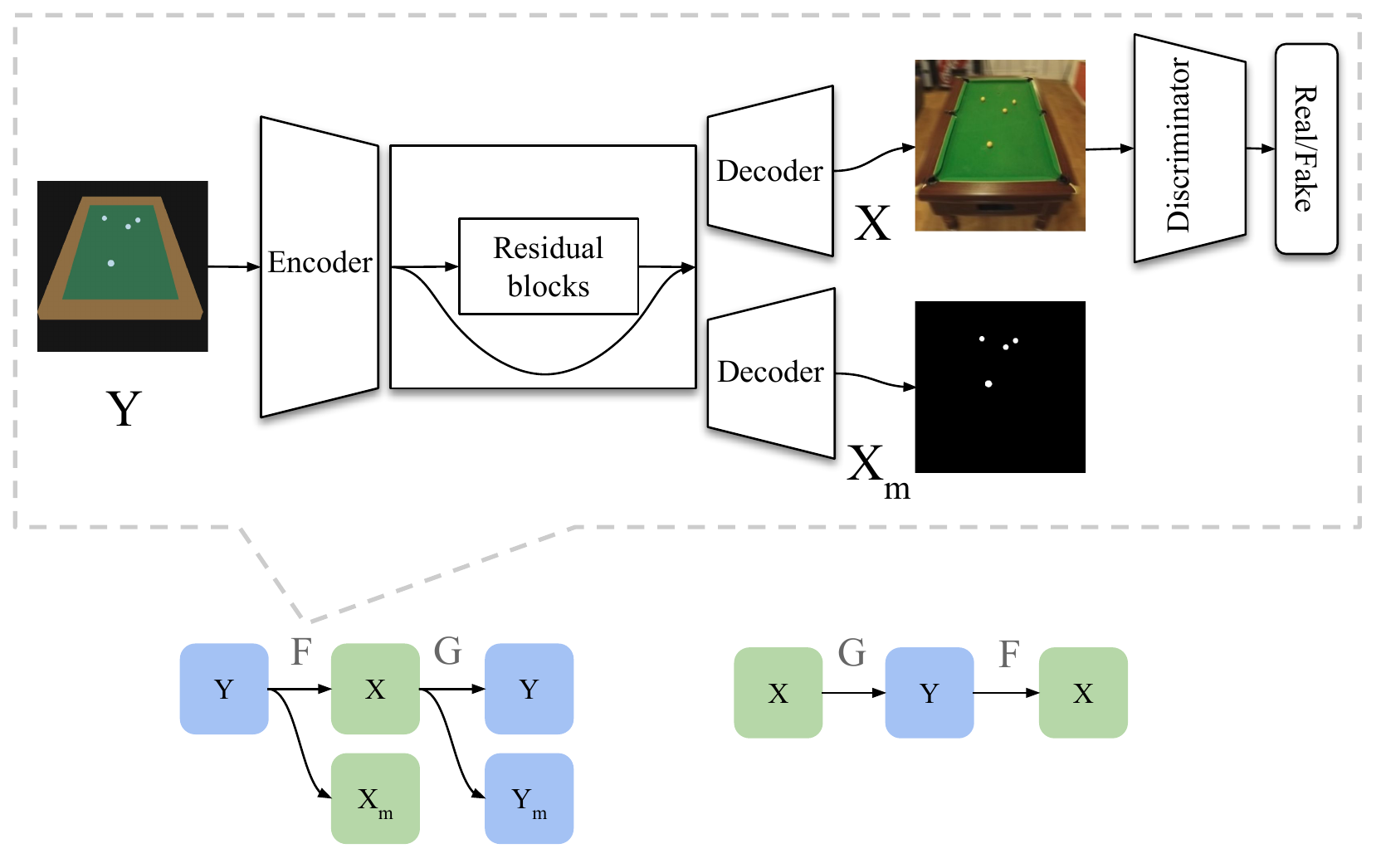}
  \caption{The domain transfer module. Bottom: mapping functions between domains. Top: detailed view of F.}
  \label{fig:domaintransfer}
\end{figure}

In order to transfer the domain between real images and synthetic images, we use unpaired images to carry out image-to-image translation.
The objective is to learn mapping functions $G: X \mapsto Y$ and $F: Y \mapsto X$ between two domains $X$ and $Y$ given two sets of unpaired training samples $\{x\} \in X$ and $\{y\} \in Y$.
Two discriminators $D_X$ and $D_Y$ classify $G(x)$ and $F(y)$ output images as real or fake by learning a perception-level representation of the inputs (\ref{fig:domaintransfer}).

A cycle-consistency loss term (\cite{cyclegan17}) is added in order to add structure to the adversarial losses $\mathcal{L}_{G}$ and $\mathcal{L}_{F}$:
\begin{equation}
\mathcal{L}_{cycle} = E_x || F(G(x))-x  ||_1 + E_y || G(F(y))-y ||_1 
\end{equation}

Similar to \cite{46717}, in order to anchor the translated images in the synthetic domain on a semantic level that preserves object position and identity, it is necessary to add auxiliary loss functions. We add auxiliary loss function $\mathcal{L}_{mask}$, but we use a different approach compared to \cite{46717}.
In particular, we extract object segmentation masks from the synthetic domain images. We let the generators output a segmentation mask in addition to the domain-adapted output, and we compute L2 loss against the ground-truth mask. As a result, the mask loss informs both generators, that share the same latent representation, enforcing the semantics of the image to be preserved (i.e., spatial position of objects).
The advantage of this approach is that semantic segmentation masks are natively available from the simulator, and are sufficient for the semantic consistency loss to be back-propagated to the whole network. 

Thus, the total domain transfer loss is:
\begin{equation}
\mathcal{L}_{DA} = \mathcal{L}_{G} + \mathcal{L}_{F} + \mathcal{L}_{cycle} + \mathcal{L}_{mask} 
\end{equation}

\subsection{Egocentric to Allocentric Viewpoint Transform} 

In order to make our framework invariant to camera perspective in the input video, we use a Spatial Transformer Network (STN) \cite{stn} architecture as a learnable image warping module in order to warp the input images to a canonical view of the scene. For the example of billiard balls, we warp the input images to a bird's-eye orthographic view of the table. This removes perspective artifacts and provides the maximum amount of information to the underlying network.

\begin{figure}[h]
  \centering
  \begin{minipage}[b]{\columnwidth}
  \includegraphics[width=\textwidth]{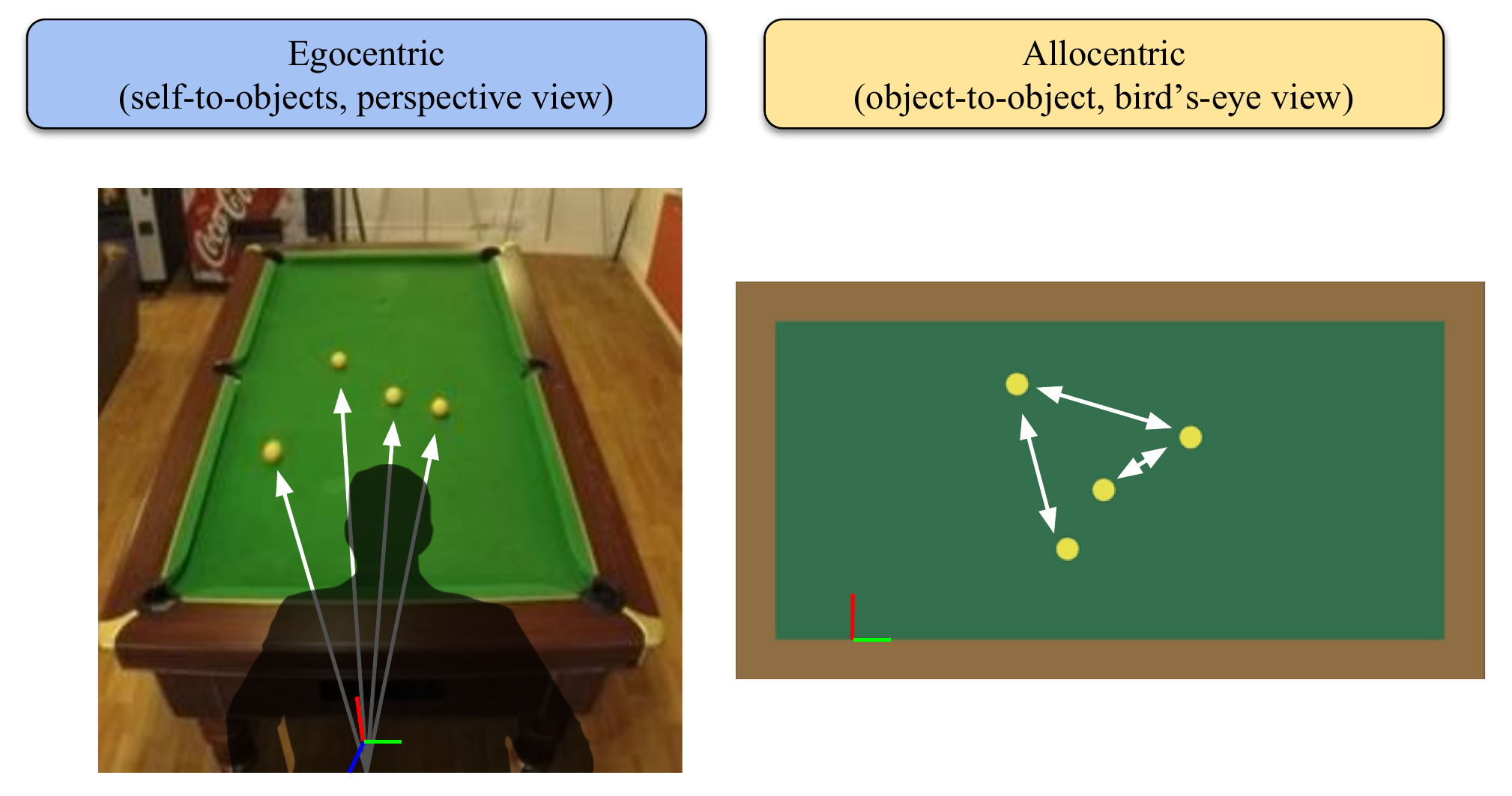}
  \caption{Spatial memory representations.}
  \label{fig:spatialmemory}
  \end{minipage}
  \begin{minipage}[b]{\columnwidth}
  \centering
    \vspace{1cm}
  \includegraphics[width=0.9\columnwidth]{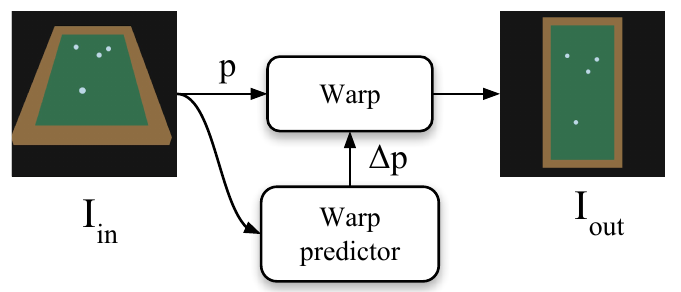}
  \caption{Egocentric to allocentric view transform module.}
  \label{fig:egocentric2allocentric}
  \end{minipage}
\end{figure}

STNs are differentiable modules that allow the spatial
manipulation of data within the network, giving neural networks the ability
to spatially transform feature maps.
The action of the spatial
transformer module is conditioned on individual data samples, with the appropriate behaviour learnt during
training for the task in question. Unlike pooling layers, where the receptive
fields are fixed and local, the spatial transformer module is a dynamic mechanism that can
actively spatially transform an image or a feature map by producing an appropriate transformation
for each input sample. The transformation is then performed on the entire feature map (non-locally)
and can include scaling, cropping, rotations, as well as non-rigid deformations. 
This allows networks, including spatial transformers, to not only select regions of an image that are most relevant (implementing an attention mechanism), but also to transform those regions to a canonical, expected view, to simplify inference in the subsequent layers. Spatial transformers can be trained with standard back-propagation,
allowing for end-to-end training of the models they are injected in.

In \cite{lin2017inverse} the structure is modified such that the network propagates warp parameters instead of warped images directly. This solves the boundary effect problem of STNs and enables a natural recurrent implementation by composing a series of warp transformations.
The warp operation can represent any transformation (e.g., affine, perspective).

A spatial transformer learns a warping $\mathbf{p}$ of an input image $\mathcal{I}$ conditioned on the image:
\begin{equation}
\mathcal{I}_{out}(\mathbf{0}) = \mathcal{I}_{in}(\mathbf{p}), \quad \mathbf{p} = f( \mathcal{I}_{in}(\mathbf{0})),
\end{equation}
where $\mathbf{0}$ is the identity warp.
In the original STN this is achieved by a \emph{localisation network} that outputs a transformation $\mathcal{T}_\theta$, a parametrised sampling grid and a differentiable image sampling module.
$f$ corresponds to a linear regressor $\mathbf{R}$ plus a bias term $\mathbf{b}$, such that:
\begin{equation}
\Delta \mathbf{p} = \mathbf{R} \cdot \mathcal{I}(\mathbf{p}) + \mathbf{b}
\end{equation}
The egocentric to allocentric transformation module is shown in Figure \ref{fig:egocentric2allocentric}.


\subsection{Recurrent Prediction of Physical Interactions}

To carry out the prediction of physical interactions, we propose the Recurrent Latent Variation Networks (RLVN). 
The model is implemented based on Convolutional LSTM networks, which extends traditional LSTMs with convolutional structures in the input-to-state and state-to-state transitions. 
Here, we decompose the recurrent model into three components: encoder, latent variation and decoder.

At each time step $t$, the encoder takes an image $x_t$ as input and produces its dense representation. 
Then, conditioned on the dense representation, a latent distribution $p(z_t|x_t)$ is constructed for capturing the physical variations $z_t$.
The decoder combines the information from the encoder via skip-connections and the latent residual to generate the predicted image as an up-convolutional decoder $p(x_{t+1}|x_t,z_t)$. 
Here, we employ variational inference to carry out the learning of the latent variable model. 
The variational lower bound can be derived as:
\begin{equation}
\begin{split}
        \log p(X) & = \sum_t \log \int p(x_{t+1}|x_t,z_t)\cdot p(z_t|x_t) d z_t \\
         & \geq \sum_t \{ \mathbb{E}_{q(z_t|x_t,x_{t+1})} [ \log  p(x_{t+1}|x_t,z_t) ] - \\ 
       & - D_{KL}[q(z_t|x_t,x_{t+1}) || p(z_t|x_t)] \}, 
\end{split}
\end{equation}
where $x_t$ is the image at $t$th time step and $z_t$ is the latent variation. 
The generative distribution is defined as a parameterised diagonal Gaussian distribution $p(z_t|x_t) = \mathcal{N}(\mu_t,\sigma_t^2)$, where $\mu_t$ and $\sigma_t$ are the parameters generated by the encoder $(\mu_t, \sigma_t) = E(x_t)$.
Similarly, the variational distribution $q(z_t|x_t, x_{t-1})$ is constructed as $\mathcal{N}(\hat{\mu}_t,\hat{\sigma}_t^2)$ in order to approximate the posterior $p(z_t|x_t, x_{t+1})$, where $\hat{\mu}_t$ and $\hat{\sigma}_t$ are generated based on both the input $x_t$ and the observation $x_{t+1}$, and $(\hat{\mu}_t,\hat{\sigma}_t) = \hat{E}(x_t, x_{t+1})$. Here, both the generative distribution $p(z_t|x_t)$ and variational distribution $q(z_t|x_t, x_{t+1})$ are jointly learned while optimising the variational lower bound.

Hence, during variational inference, we apply the reparameterisation trick \cite{kingma2013auto} by sampling $\epsilon \sim \mathcal{N}(0,I)$ which yields to $\hat{z}_t=\hat{\mu_t} + \epsilon\cdot \hat{\sigma_t}$. The estimated lower bound can be derived as:
\begin{equation}
\begin{split}
        \hat{\mathcal{L}} & \approx  \sum_t \{ \log  p(x_{t+1}|x_t,\hat{z}_t) \\
        & - D_{KL}[q(z_t|x_t,x_{t+1}) || p(z_t|x_t)] \},
\end{split}        
\end{equation}
where the Kullback-Leibler divergence term is integrated as a Gaussian KLD, and the gradients can be directly constructed and back-propagated through the neural network.

\begin{figure}[t]
  \centering
	\includegraphics[width=\columnwidth]{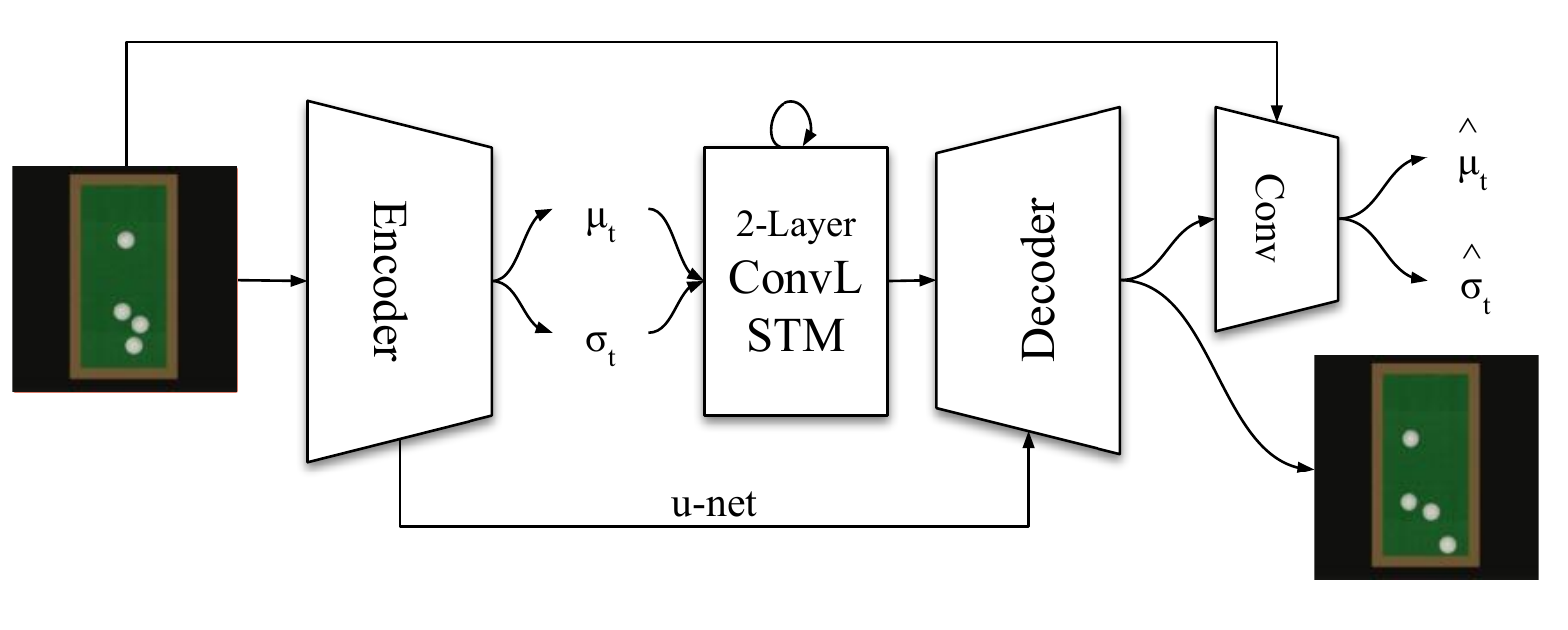}
  \caption{The RLVN physics predictor module.}
  \label{fig:predictor}
\end{figure}

The combination of variational inference and u-net shape allows the network to learn expressive representations of the scenes.
Intuitively, by decomposing the latent variation $z_t$ from the network, we place an inductive bias into the model to separate the learning of the scene/object appearance and the dynamic interactions (including the position, velocity, mass and friction components of the objects).
In this way, the u-net (as shown in Figure \ref{fig:predictor}) is encouraged to learn the scene/object appearance and construct deterministic representation, while the latent variation attempts to capture the dynamic interactions which is considered as stochastic representation. 
Therefore, the decoder is able to easily construct the predicted images by combining the two representations. 

Compared to the deterministic counterpart which has no latent variation (i.e., $z_t$ is not drawn from a latent distribution, but is directly generated from the encoder instead), our framework has better capacity for predicting complex dynamic physical interactions. In addition, the deterministic model overfits the dataset very quickly. Interestingly, it gradually ignores the prediction of dynamic interactions but only focuses on learning the scene/object appearance.

\begin{table}[t]
\centering
\resizebox{0.75\columnwidth}{!}{
\begin{tabular}{l}
Domain transfer module  \\
\hline \\
$\lbrack$ Encoder layer 1 $\rbrack$ Conv. $7^2$, Stride $1^2$, ReLU activ.\\
$\lbrack$ Encoder layer 2-3 $\rbrack$ Conv. $3^2$, Stride $2^2$, ReLU activ.\\
$\lbrack$ Encoder layer 4-9 $\rbrack$ Resid. $3^2$ \\
\\
$\lbrack$ Decoder layer 10-11 $\rbrack$ Upconv. $3^2$, Stride $0.5^2$, ReLU activ.\\
$\lbrack$ Decoder layer 12 $\rbrack$ Conv. $7^2$, Stride $1^2$, ReLU activ.\\
$\lbrack$ Dis layer 1-4 $\rbrack$ Conv. $3^2$, Stride $2^2$, ReLU activ.\\
\\

View transform module (x4 recursion) \\
\hline \\
$\lbrack$ Layer 1-2 $\rbrack$ Conv. $7^2$, Stride $1^2$, ReLU activ.\\
$\lbrack$ Layer 3 $\rbrack$ FC 48\\
$\lbrack$ Layer 4 $\rbrack$ FC 8\\
$\lbrack$ Layer 5 $\rbrack$ Warp op. $\rightarrow \lbrack$ layer 1$\rbrack$ \\
\\

RVLN predictor \\
\hline \\
$\lbrack$ Encoder layer 1 $\rbrack$ Conv. $4^2$, Stride $2^2$, ReLU activ.\\
$\lbrack$ Encoder layer 2 $\rbrack$ Conv. $4^2$, Stride $1^2$, ReLU activ.\\
$\cdots$      \\
$\lbrack$ Encoder layer 7 $\rbrack$ FC 1000\\
$\lbrack$ Encoder layer 8 $\rbrack$ $\mu$ FC 400, $\sigma$ FC 400\\
$\lbrack$ Encoder layer 9 $\rbrack$ FC 2048\\
$\lbrack$ Encoder layer 10 $\rbrack$ ConvLSTM\\
\\
$\lbrack$ Decoder layer 10 $\rbrack$ Deconv. $4^2$, Stride $1^2$ $[$CONCAT$]$\\
$\lbrack$ Decoder layer 11 $\rbrack$ Deconv. $4^2$, Stride $2^2$ $[$CONCAT$]$\\
$\cdots$      \\
$\lbrack$ Conv layer 1-8 $\rbrack$ Same as Encoder 1-8\\

\end{tabular}
}
\caption{Implementation details for network modules.}
\label{tab:networkparameters}
\end{table}

\section{Experimental Results}
We first validate our proposed physics predictor by comparing it to different state-of-the-art baselines on synthetic billiard videos. The billiard scenario is ideal to evaluate long-term prediction, since it is a chaotic system; even non-skilled humans have difficulty in making medium-term predictions.
Then we test the complete framework in the actual complex real scenario: real billiard videos with multiple camera positions.
The network was implemented using Tensorflow. The implementation details are reported in Table \ref{tab:networkparameters}.

\textbf{Data generation}
Simplified 2D billiard-like bouncing balls scenarios are a common benchmark for physics prediction \cite{Fragkiadaki16,van2018relational,Bhattacharyya2018LongTermIB}.
For our experiments we generate a more realistic 3D dataset using Blender, with Bullet as the underlying physics engine.
In each video four balls of similar mass and size are placed at random and with random velocities on a billiard table. The balls and billiard table behave in a realistic way, including friction and restitution forces.
Each video is composed by 20 frames. We train the network on the first 10 frames and predict the successive 10, evaluating the predictions against the true frames.
We generate 10k video episodes for training and 1k for testing.

We then capture several billiard videos from different camera angles and different lightning conditions, with four balls of the same color.
The videos are captured from three different viewpoints, and the video sequences are manually cut in order to remove players occluding the image. The longer video sequences are then segmented into sequences of 20 frames.

\begin{figure*}[h!]
  \centering
	\includegraphics[width=\textwidth]{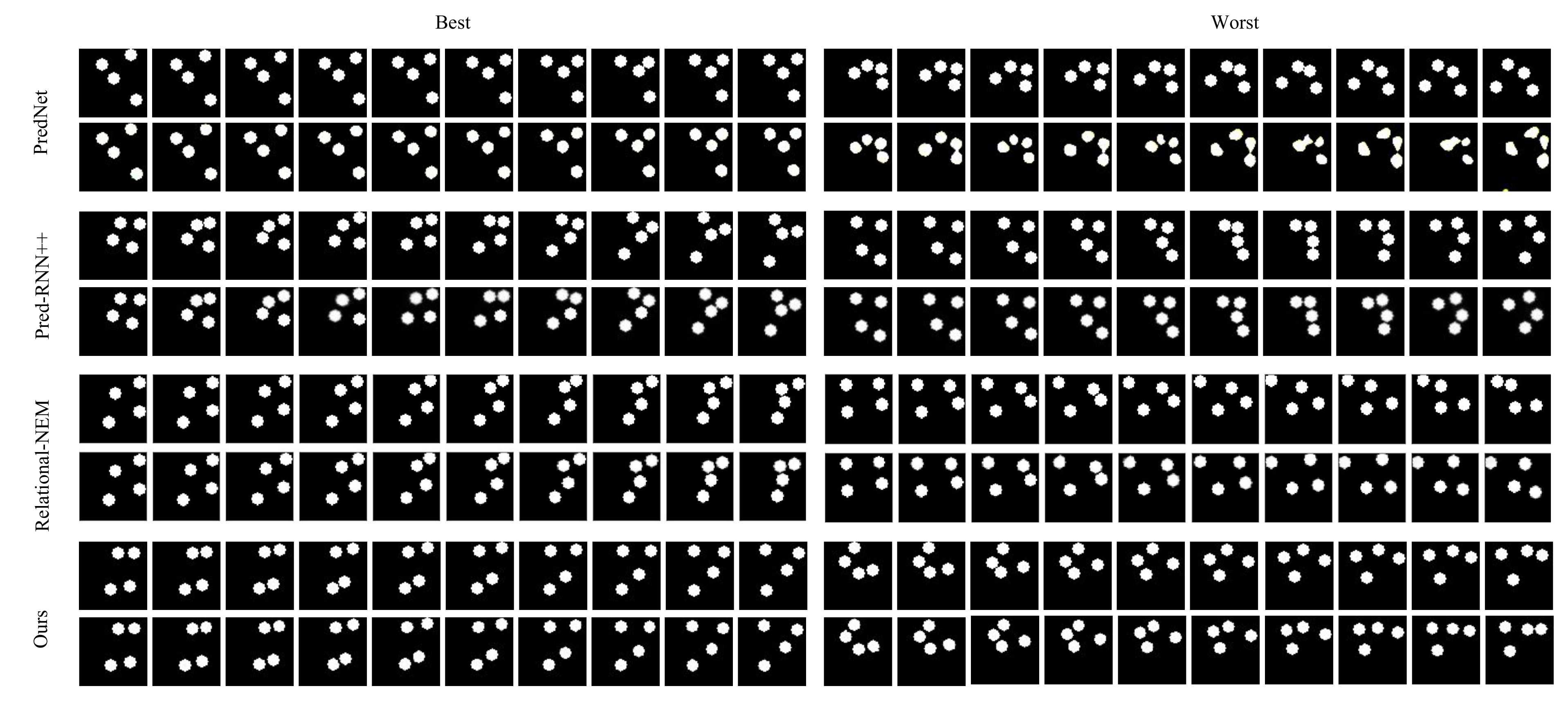}
  \caption{Comparison of our approach to PredNet, R-NEM and Pred-RNN++ over 10 prediction steps. For each method the first line shows the ground-truth and the second line shows the predicted sequence. All networks were fed a sequence of 10 steps in order to predict the next 10.
Sequences on the left: best results; sequences on the right: worst results.}
  \label{fig:exp1}
\end{figure*}

\subsection{Validating the physics predictor}
We explore how our predictor learns the physical object
state (position, velocity, mass, and friction) and we compare the ability of our predictor to predict long sequences to three recent state-of-the-art baselines (PredNet \cite{prednet16}, Relational-NEM \cite{van2018relational} and Pred-RNN++ \cite{wang2018predrnn}) on the synthetic billiard dataset.
All networks are trained on 64$\times$64 grayscale images, with a batch size of 8, with the exception of Relational-NEM, which was trained with binarized inputs as in the original implementation.
Given a sequence of 10 frames, we predict the next 10, and compare against the ground-truth.
PredNet was trained for next frame prediction on sequences of 20 frames and successively fine-tuned on full sequences of 20 frames.

\begin{figure}[h!]
  \centering
  \includegraphics[width=\columnwidth]{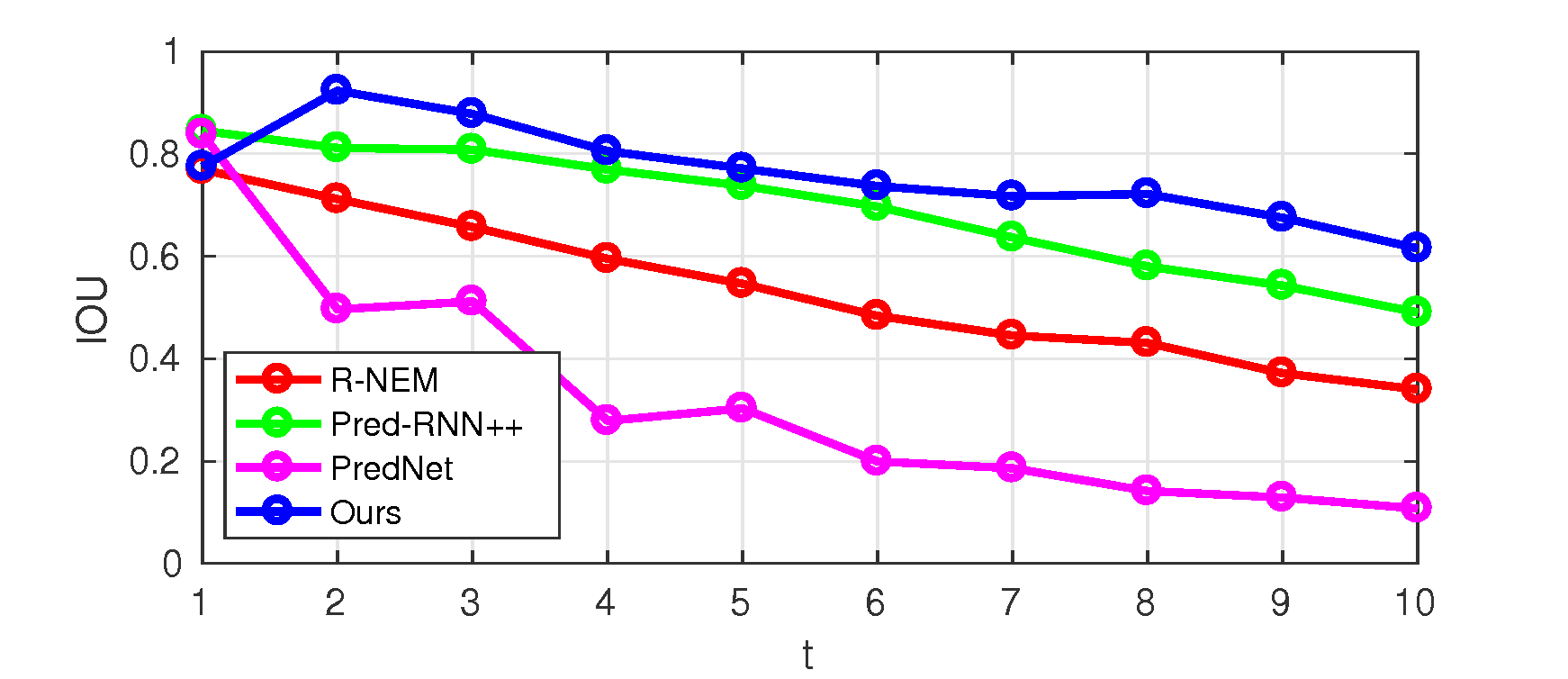} \\
  \includegraphics[width=\columnwidth]{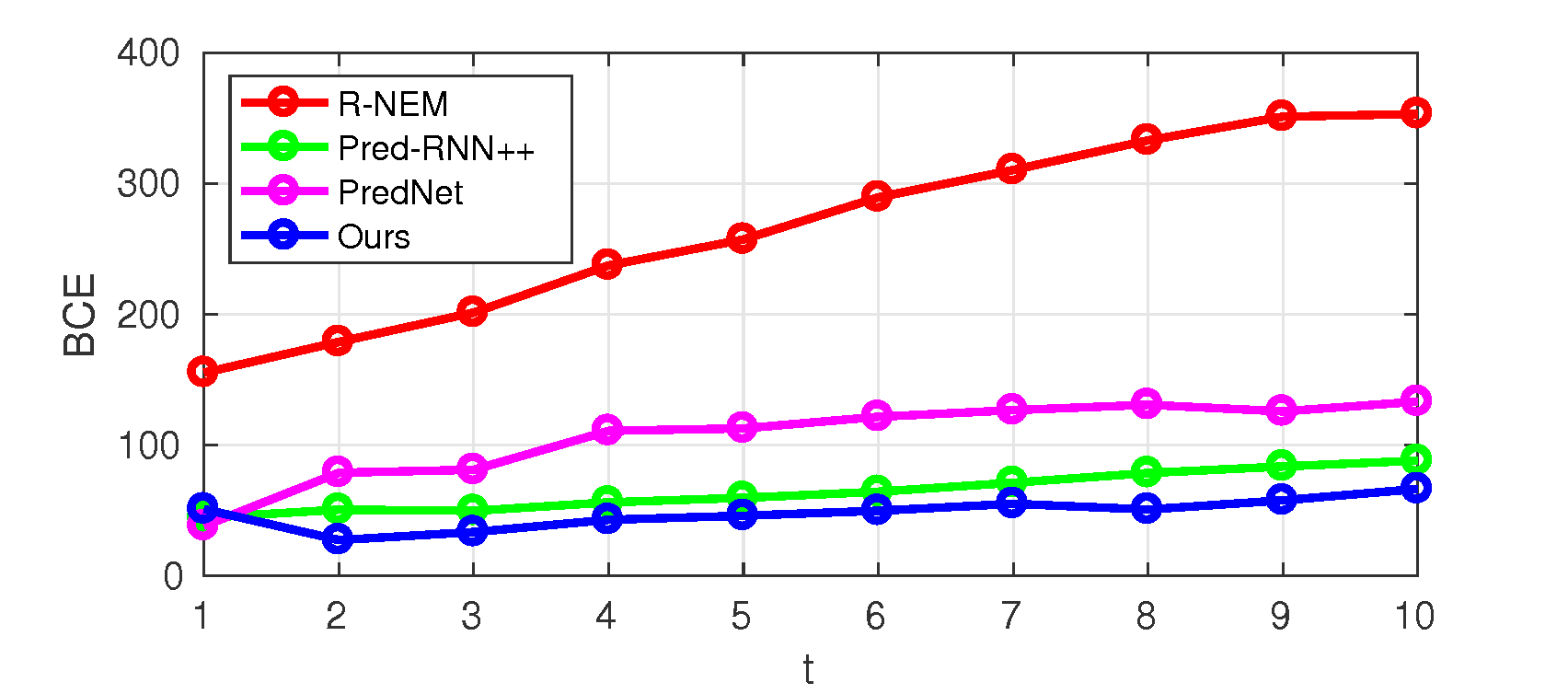} 
  \caption{Average IOU and BCE scores for the four approaches over 10 prediction steps.}
  \label{fig:exp1norm}
\end{figure}

Figure \ref{fig:exp1} shows best and worst qualitative results on random test sequences, while Figure \ref{fig:exp1norm} reports the \emph{Intersection Over Union} (IOU) and \emph{Binary Cross-Entropy} (BCE) scores for the four approaches over 10 future predicted time steps. 


IOU score is defined as:
$$
IOU = { { \sum_{ij} \left[ \mathcal{I}(Y_{ij} > p) * \mathcal{I}(X_{ij}) \right]} 
\over { \sum_{ij} \left[  \mathcal{I}(\mathcal{I}(Y_{ij} > p) + \mathcal{I}(X_{ij})) \right]} }
$$
where $\mathcal{I}$ is the indicator function, $Y_{ij}$ is the predicted value at position $(i,j)$, $X_{ij}$ is the true value $(i,j)$, and $(i,j)=1 \text{ if } (i,j)>p$. In our experiments, $p$ is set to 0.8. 

It can be seen that PredNet is able to predict future frames with reasonable accuracy up to 4-5 steps in the future. This is in line with the results from \cite{prednet16}.
Relational-NEM generates accurate predictions for most sequences, while not being able to correctly predict in a few cases. This was due to the network not being able to learn disentangled representations for the four balls in a few cases. 
Pred-RNN++ is able to generate accurate predictions for most sequences on our dataset.
As expected, the prediction accuracy of all methods decreases over time, with PredNet showing increased degradation.

Our method shows consistently more accurate predictions over time for all sequences compared to competing approaches, with an IOU of and comparable accuracy degradation over time to Pred-RNN++.

\begin{figure*}[t]
  \centering
  \includegraphics[width=0.95\textwidth]{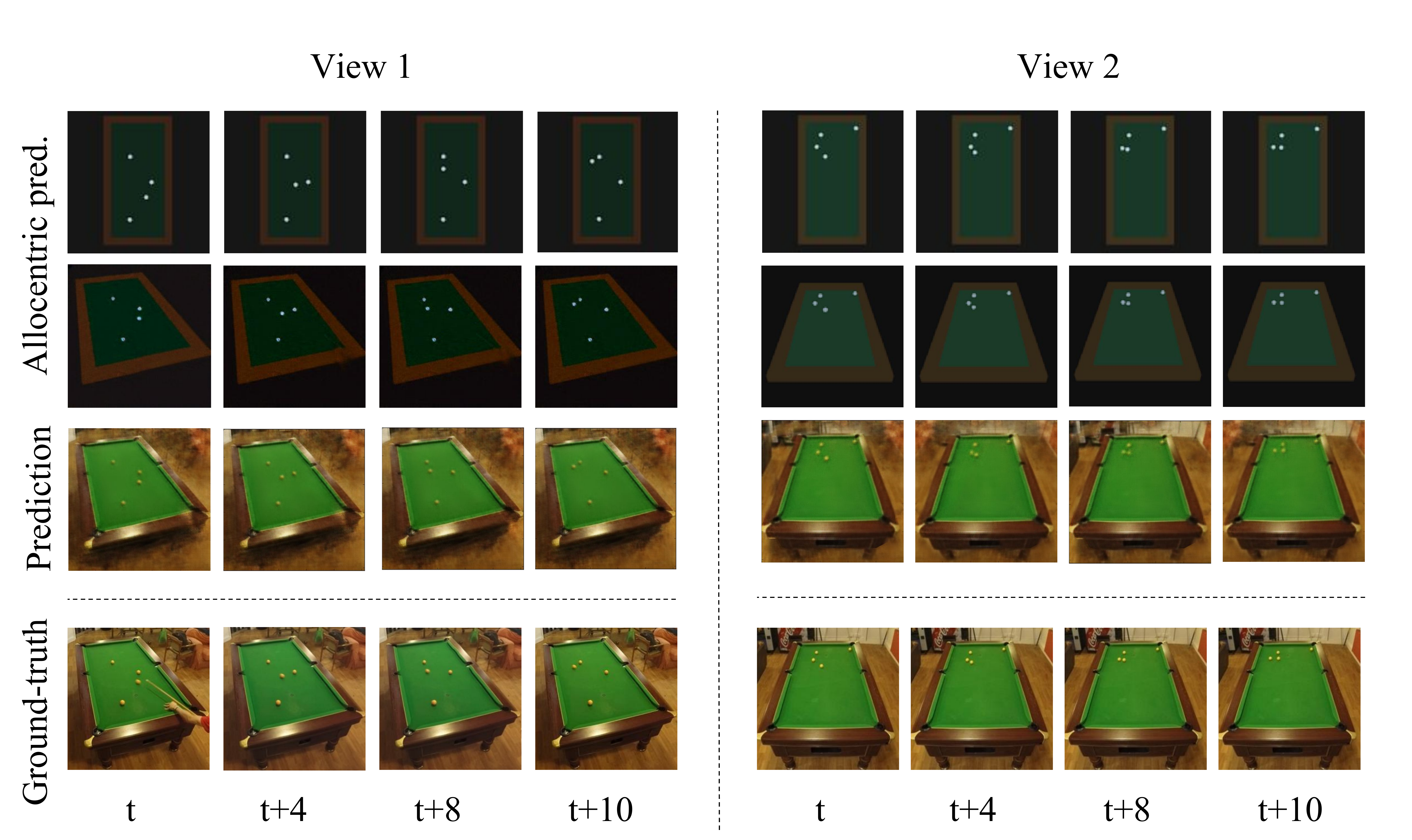}
  \caption{Results for the multi-view camera scenario. Given a sequence of 10 input frames at steps, the network predicts 10 frames in the future. First row: predicted allocentric view; second row: predicted view in the original egocentric frame; third row: final prediction in the real domain; last row: ground truth. First column: last input frame; following three columns: predicted time steps.}
  \label{fig:exp2}
\end{figure*}

\subsection{Billiard Tables with Multiple Camera Views}
We now evaluate the complete framework on a series of real videos from different viewpoints. The camera remains static for the duration of each video sequence.
We first train the domain transfer module on an unpaired training set of 40k real and synthetic samples.
We then train the allocentric viewpoint transform module on a synthetic dataset of 20k samples.
Finally, the physics predictor is trained on a synthetic dataset of 10k video sequences, for a total of 200k samples.

\begin{figure}[h!]
  \centering
	\includegraphics[width=0.8\columnwidth]{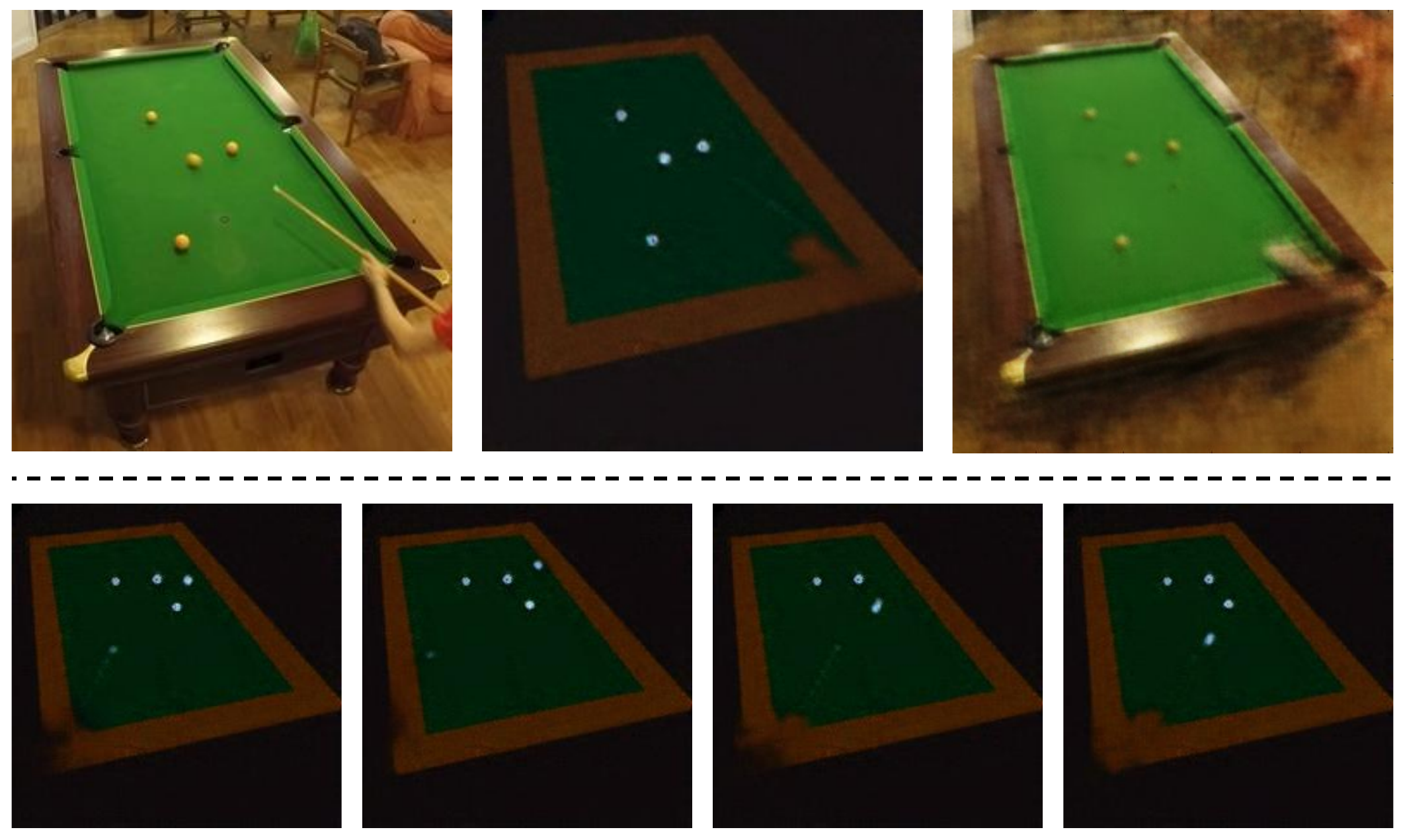}
  \caption{Examples of failed predictions due to foreign objects. Top: a hand with cue stick (top-left) causes allucinated object in the synthetic and real predicted images (top-middle and top-right). Bottom:the point of the cue stick is incorrectly recognized as a ball and causes the prediction to fail.}
  \label{fig:failures}
\end{figure}

In Figure \ref{fig:exp2} we show qualitative results on sequences with different out-of-sample viewpoints. It is possible to see how the images are transported to the allocentric representation and the predictions are then transported back to realistic images.
It should be noted that the presence of occlusions (e.g., the player's hand occluding part of the table) can have a negative effect on the predictions by propagating through the network. 
Figure \ref{fig:failures} shows some of the effects of failures in the style transfer module on the predicted images.

\section{Conclusion}
Generative models for intuitive physics understanding are showing promising performances, but they are still mostly limited to toy examples and require many thousands of examples in order to gain the ability of transferring to real world scenario or to deal with different viewpoints.
We proposed a neuro-inspired framework that can learn from fewer examples, by projecting arbitrary viewpoints to a canonical ensemble view of the scene and to a canonical image domain.
The domain transformation from real images to canonical images is possible by training the networks on unpaired image sets.
Meanwhile, the projection from arbitrary to canonical views can be trained with pairs of synthetic images only, making the module independent from labeled real data.
In addition, we proposed RLVN, a novel physical predictor which decomposes the learning of scene representation and objects' dynamics, which grants strong ability to carry out predictions on physical dynamics in different scenarios. 
By decomposing the latent variation from the network, the former is free to learn the stochastic physical properties of the objects, thus their interactions, while the latter is encouraged to learn the deterministic object and scene appearance.
The next natural step would be to investigate a two-step approach in which the architecture is bootstrapped on synthetic data and then trained by observing real data in an unsupervised manner. 



\clearpage
\small
\bibliographystyle{aaai}
\bibliography{nips2018}

\end{document}